%% file: acl_latex.tex
\pdfoutput=1
\documentclass[11pt]{article}

\usepackage[preprint]{acl}

\usepackage{times}
\usepackage{latexsym}

\usepackage[T1]{fontenc}
\usepackage[utf8]{inputenc}

\usepackage{microtype}
\usepackage{hyperref}       % hyperlinks
\usepackage{url}            % simple URL typesetting
\usepackage{booktabs}       % professional-quality tables
\usepackage{amsfonts}       % blackboard math symbols
\usepackage{nicefrac}       % compact symbols for 1/2, etc.
\usepackage{microtype}      % microtypography
\usepackage{xcolor} 
\usepackage{graphicx}% colors
\usepackage{multicol}
\usepackage{multirow}
\usepackage{makecell}
\usepackage{enumitem}

\usepackage{xcolor}         % colors
\usepackage{color, colortbl}
\definecolor{citecolor}{HTML}{2980b9}
\definecolor{linkcolor}{HTML}{c0392b}
\definecolor{darkorange}{HTML}{FF8C00}
\definecolor{chocolate}{HTML}{D2691E}
\definecolor{darkgreen}{HTML}{006400}
\definecolor{darkblue}{HTML}{00008B}
\definecolor{mediumblue}{HTML}{0000CD}
\definecolor{dodgerblue}{HTML}{1E90FF}
\definecolor{royalblue}{HTML}{4169E1}
\definecolor{shadecolor}{RGB}{237,237,237}
\definecolor{backred}{RGB}{255, 190, 190}
\definecolor{backblue}{RGB}{210, 230, 250}

\definecolor{zrrgreen}{HTML}{008000}
\definecolor{zrrblue}{HTML}{4682B4}
\definecolor{zrrred}{HTML}{B22222}

\usepackage{graphicx}
\usepackage{subcaption}
\usepackage{float}

\newcommand{\ignore}[1]{}

\newcommand{\xhdr}[1]{{\noindent\bfseries #1}.}

\usepackage{inconsolata}

\usepackage{graphicx}

\title{\textsc{MM-MATH}: Advancing Multimodal Math Evaluation with \\Process Evaluation and Fine-grained Classification}

\author{Kai Sun$^{*},$ Yushi Bai$^{*}$, Ji Qi, Lei Hou, Juanzi Li \\
Tsinghua University}

\begin{document}
\maketitle

\renewcommand{\thefootnote}{\fnsymbol{footnote}}
    \footnotetext[1]{Equal contribution
    }
\renewcommand{\thefootnote}{\arabic{footnote}}

\begin{abstract}

To advance the evaluation of multimodal math reasoning in large multimodal models (LMMs), this paper introduces a novel benchmark, MM-MATH. MM-MATH consists of 5,929 open-ended middle school math problems with visual contexts, with fine-grained classification across difficulty, grade level, and knowledge points. 
Unlike existing benchmarks relying on binary answer comparison, MM-MATH incorporates both outcome and process evaluations. Process evaluation employs LMM-as-a-judge to automatically analyze solution steps, identifying and categorizing errors into specific error types.
Extensive evaluation of ten models on MM-MATH reveals significant challenges for existing LMMs, highlighting their limited utilization of visual information and struggles with higher-difficulty problems. 
The best-performing model achieves only 31\% accuracy on MM-MATH, compared to 82\% for humans. This highlights the challenging nature of our benchmark for existing models and the significant gap between the multimodal reasoning capabilities of current models and humans.
Our process evaluation reveals that diagram misinterpretation is the most common error, accounting for more than half of the total error cases, underscoring the need for improved image comprehension in multimodal reasoning.
The code and dataset are available at \url{https://github.com/kge-sun/MM-Math}.

\end{abstract}

\input{text/001introduction}

\input{text/002datasets}
\input{text/003evaluation}

\input{text/004related_work}
\input{text/005conclusion}

\bibliography{neurips}

\newpage
\appendix
\onecolumn

\input{text/006appendix}

\end{document}

%% file: text/001introduction.tex
\section{Introduction}

% \ys{1. Background: Multimodal math evaluation of MLLM}
\begin{figure*}[!t]
    \centering
    \includegraphics[width=\linewidth]{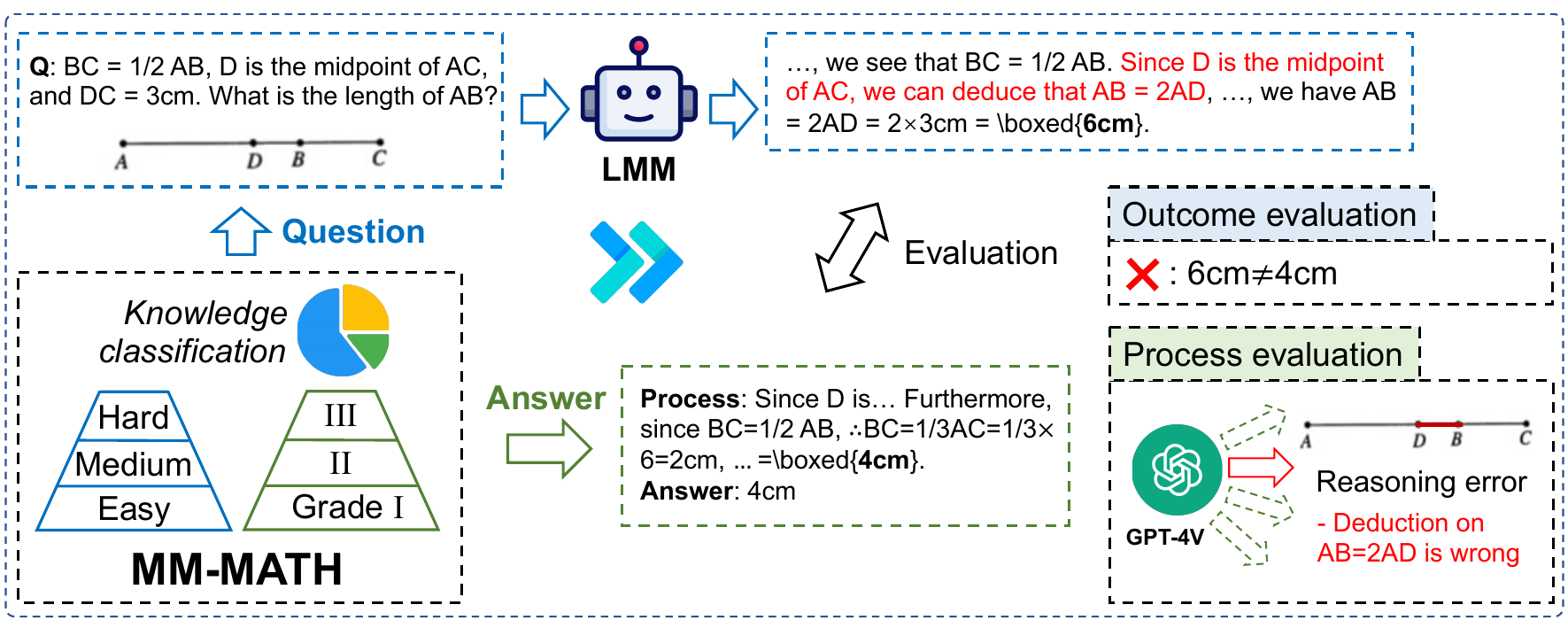}
    %\vspace{-2mm}
    \caption{An overview of the MM-MATH benchmark design. The problems are classified along their difficulty, grade level, and knowledge point.
    We include both outcome evaluation and process evaluation to identify and attribute the error in model's reasoning process.}
    %\vspace{-2mm}
    \label{fig:overview}
\end{figure*}

Due to their exceptional performance in handling complex text and images, large multimodal models (LMMs) such as GPT-4V~\cite{openai2023gpt4v} and Claude-3~\cite{anthropic2024claude3} have garnered significant interest in both industry and academia. 
Previous studies suggest that they still underperform on multimodal math reasoning tasks~\cite{chen2021geoqa,lu2023mathvista,zhang2024mathverse}, as such tasks require understanding multimodal information and interleaving reasoning within this information~\cite{lightman2023let}.
To further advance LMM's mathematical capabilities, we believe the following two issues urgently need addressing: \textbf{(1)} \textbf{What are the specific reasons that lead to the model's mistakes}, such as misunderstanding the diagram or errors in reasoning? \textbf{(2)} \textbf{How does the model perform across different categories of multimodal math problems}, 
% including various knowledge points and difficulty levels, 
and which specific types of problems does the model excel at or struggle with?

In this paper, we introduce \textbf{MM-MATH} benchmark to provide a more fine-grained and reliable assessment of LMMs' multimodal math capability.
MM-MATH comprises a total of 5,793 open-ended multimodal math problems from middle school.
We show an overview of the design of MM-MATH in Figure~\ref{fig:overview}.
To address the aforementioned issue (1), MM-MATH combines traditional outcome evaluation (comparing the model's answer to groundtruth and reaching binary result) with \textbf{process evaluation}.
Process evaluation involves using LMM-as-a-judge~\cite{zheng2024judging} to automatically identify errors in the model's output process and categorize the causes of these errors.
Concretely, we employ GPT-4V~\cite{openai2023gpt4v} to compare the step-by-step solution generated by the model with our annotated groundtruth solution, and identify the first error in the model's process to determine the main reason that leads to a wrong answer.
We categorize the causes of LMM's errors into four types, including \emph{diagram misinterpretation}, \emph{reasoning error}, \emph{calculation error}, and \emph{textual condition misunderstanding}.

In response to issue (2), MM-MATH includes \textbf{fine-grained classification}, where the problems are classified along three dimensions: difficulty, grade level, and knowledge points, to evaluate the breadth, depth, and specific knowledge for math reasoning capabilities of LMMs. 
For difficulty, we classify problems into three levels---easy, medium, and hard---based on the accuracy of human students on the problems. For grade level, we include problems in middle school, encompassing all relevant visual math problems taught in each grade. For knowledge points, each problem is classified according to a predefined three-level knowledge taxonomy by experienced teachers. These comprehensive annotations in the MM-MATH dataset result in clear difficulty distinction, extensive data coverage, and systematic knowledge organization.

We conduct an extensive evaluation of both open-source and closed-source LMMs on MM-MATH. 
Outcome evaluation reveals that our benchmark poses significant challenges for existing LMMs. For example, the latest Sota model, GPT-4o~\cite{gpt4o}, achieves an accuracy of only 31\%, compared to an 82\% accuracy of human students. 
Moreover, all models perform poorly on hard-level problems, with none exceeding 11\% accuracy, and some models even fail to solve any problems correctly.
We further find that current LMMs' multimodal reasoning remains primarily text-based, lacking effective utilization of graphical information. 
This is evidenced by the minimal accuracy difference---only 2-3 percentage points---between when the model is given only textual input and when it is provided with both text and images.
Our process evaluations show that diagram misinterpretation accounts for more than 50\% of the total errors for current LMMs, suggesting the most critical direction for improvement is enhancing their abilities to recognize and interpret math diagrams.

%% file: text/002datasets.tex
\section{MM-MATH}
\label{datasets}

\begin{table*}[htbp]
  \centering
  \caption{Comparison of our MM-MATH benchmark with existing multimodal benchmarks. For the `size' column, we only include the number of multimodal math problems in each benchmark.}
  \resizebox{\textwidth}{!}{ 
    \begin{tabular}{llllcc}
    \toprule
    \textbf{Benchmark} & \textbf{Size} & \textbf{Question Type} & \textbf{Grade} & \textbf{Fine-grained Classification} & \textbf{Process Evaluation} \\
    \midrule
    UniGeo~\cite{chen2022unigeo} & 4,998  & choice & middle school &   \checkmark &  \\
    GeoQA~\cite{chen2021geoqa} & 5,010  & choice & middle school \\
    GeoQA+~\cite{cao2022augmented} & 2,518  & choice & middle school \\
    Geometry3K~\cite{lu2021inter} & 3,002 & choice & middle school \\
    OlympiadBench~\cite{he2024olympiadbench} & 3,102 & open-ended & Olympiad-level & \checkmark \\
    MathVista~\cite{lu2023mathvista} & 6,141 & choice \& open-ended & - \\
    MathVerse~\cite{zhang2024mathverse} & 2,612  & choice \& open-ended & - &   & \checkmark \\
    \midrule
    MM-MATH & 5,929 & open-ended  & middle school & \checkmark & \checkmark \\
    \bottomrule
    \end{tabular}%
    }
  \label{tab:cmp}%
\end{table*}%

\begin{table}[!ht]
\setlength{\tabcolsep}{4pt}
    \centering
    \caption{Key statistics of MM-MATH.}
    \label{tab:CARP}
    \resizebox{0.35\textwidth}{!}{
    \begin{tabular}{lr}
    \toprule
    \textbf{Statistic} & \textbf{Number} \\
    \midrule
    Total Problems & 5,929 \\
    \midrule
    \textbf{Difficulty} &  \\
    \hspace{3mm}*Easy  & 378 \\
    \hspace{3mm}*Medium & 4,488 \\
    \hspace{3mm}*Hard  & 1,063 \\
    \midrule
    \textbf{Grade} &  \\
    \hspace{3mm}*Grade Seven & 682 \\
    \hspace{3mm}*Grade Eight & 2,590 \\
    \hspace{3mm}*Grade Nine  & 2,657 \\
    \midrule
    Average Question Length & 488 \\
    Average Answer Length & 275 \\
    Max Question Length & 2,391 \\
    Max Answer Length & 2,781 \\
    \bottomrule
    \end{tabular}%
    }
\end{table}

\subsection{Overview of MM-MATH}
\xhdr{Design Principle} Multimodal mathematical reasoning tasks demand an understanding of both the problem's text and the associated diagram, requiring math reasoning to produce a step-by-step solution that leads to the final answer. 
% In secondary mathematics education, connecting mathematical expressions with corresponding diagrams is a crucial ability for students, exemplified by linking parabolas on a coordinate axis with quadratic equations. 
% Base on this, we construct a graph-based mathematical reasoning dataset comprised of real-world secondary school examination questions to assess math solving abilities. 
We adopt an open-ended format for two reasons: 1) Other formats, such as multiple choice, make it easier for the model to guess the correct answer by chance~\cite{wang2024mmlu}. 2) Open-ended format better facilitates step-by-step solution process to help identify the error in the model's response.
We adhere to the following principles when constructing MM-MATH:

\begin{itemize}[itemsep=0pt, leftmargin=*]
    \item \textbf{Comprehensive coverage}: We aim to cover as many types and difficulty levels of problems as possible. Consequently, we collect all math problems that contain visual content from exams and textbooks used in secondary schools.
    \item \textbf{Computation problems only}: While math problems may include proofs, computations, and drawings, we exclusively select computation-type problems for our dataset. 
    \item \textbf{Uniform data format}: Each problem in the dataset includes a question statement, an image, a human-annotated step-by-step solution process, and multi-dimensional metadata annotations.
    \item \textbf{Multi-dimensional metadata annotations}: For each problem, we also provide its grade level, difficulty, and knowledge point tagging from human educational taxonomy as its metadata.
    % \begin{itemize}
    %     \item \textbf{Grade-Level Categorization}: Grade level represents the knowledge acquired by students over a specific period, typically divided by academic year in real-world education.
    %     \item \textbf{Difficulty Classification}: Difficulty level indicates the complexity of a problem, usually determined by the average score obtained in exams, with difficulty coefficients assigned accordingly.
    %     \item \textbf{Knowledge Point Tagging}: Knowledge point refers to the primary or secondary theorems utilized in each problem, providing crucial insights for problem-solving approaches.
    % \end{itemize}
\end{itemize}

\begin{figure}[t]
    \centering
    \includegraphics[width=0.8\linewidth]{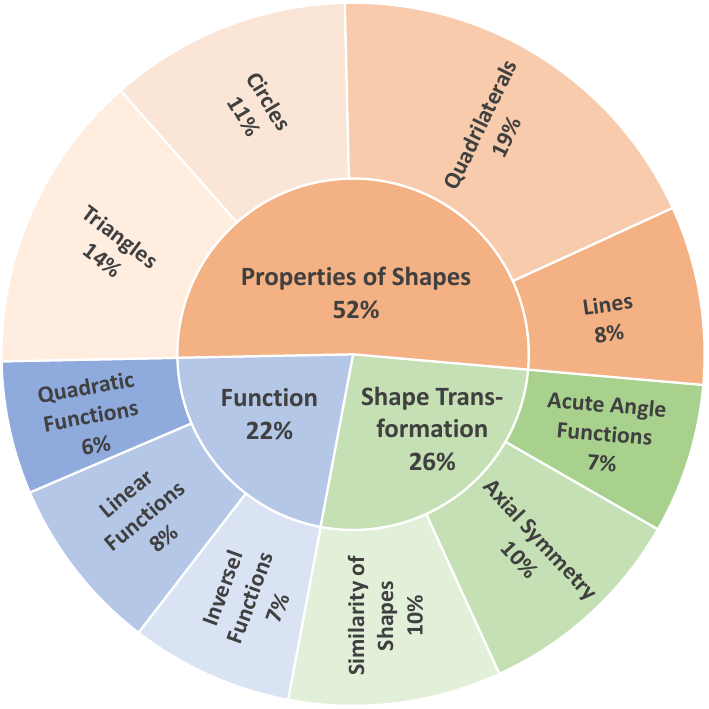}
    \caption{Knowledge point distribution of MM-MATH. \emph{Properties of Shapes} refers to the characteristics of different shapes, \emph{Shape transformation} investigates the deformation and movements of shapes, and \emph{Function} refers to the mutual reasoning between algebraic expressions and graphs.}
    \label{fig:process_pie}
\end{figure}

\xhdr{Dataset Overview}
MM-MATH is the first multimodal math benchmark to include process evaluation and fine-grained classification, as highlighted in the comparison of existing multimodal benchmarks in Table~\ref{tab:cmp}.
Detailed statistics for MM-MATH are provided in Table~\ref{tab:CARP}, and the distribution of knowledge points is illustrated in the pie chart in Figure~\ref{fig:process_pie}.

\subsection{Dataset Construction Pipeline}

\xhdr{Data collection}
The problems in MM-MATH dataset are sourced from the \href{https://www.21cnjy.com/}{21st Century Education Network}, which is one of the largest online question banks for primary and secondary schools in China. It provides a comprehensive collection of challenging, curriculum-aligned, and exam-relevant questions designed to assess student learning capabilities. We restrict the problems from the 2021-2022 academic year, manually filtering for computational math problems with visual context. 
% These selected questions are then extracted by the website engineers.

\xhdr{Format transformation} 
Most of the problems from the original database are in MathML format. However, considering the widespread use of LaTeX in existing mathematical datasets, we devise a systematic approach to convert MathML into standard LaTeX format for easier integration with other datasets. 
Specifically, we utilize MathConverter\footnote{\href{https://github.com/hexinnovation/MathConverter}{https://github.com/hexinnovation/MathConverter}.} to transform MathML representations of mathematical formulas into LaTeX. For instance, \(\frac{1}{2}\) in MathML is converted to \textbackslash frac\{1\}\{2\} in LaTeX. 
Additionally, we establish string conversion rules to change symbol elements into LaTeX format.
For example, we convert ``\textbackslash text\{$\triangle$\}'' to ``\textbackslash triangle''.
To address the use of non-standard punctuation in Chinese strings, such as full-width plus signs,  we leverage GPT-4~\cite{achiam2023gpt} for conversion, with manual verification of the final output. 
This systematic process ensures the accuracy of numerical values in LaTeX while maintaining readability and standardization in the output. 
During GPT-4 processing, we also encapsulate the final answers within \textbackslash boxed\{\}, 
a technique inspired by the construction approach of the MATH dataset~\cite{hendrycks2021measuring}, which facilitates comparison with groundtruth answers for outcome evaluation.

Our collected data contains four distinct question types: multiple-choice, fill-in-the-blank, open-ended, and composite questions. 
% Aside from standard answers, step-by-step solutions are also provided. To facilitate the detection of intermediate errors during comparison evaluation, all question types are uniformly converted into open-ended questions. 
We convert them into uniform open-ended questions in the following manners. 
For multiple-choice and fill-in-the-blank questions, we rephrase them into open-ended forms and extract their explanations as step-by-step derivations. 
For composite questions with a common textual problem and multiple sub-questions, we treat each statement as the premise for sub-questions, integrating the conclusions of preceding sub-questions into the subsequent ones. 
More details for the transformation process are presented in Appendix~\ref{open_end transformation}.

Additionally, since the original data is in Chinese, catering to Chinese students, we translate the dataset into English using GPT-4. To ensure accuracy, we manually verify the translations. This effort aims for a fairer comparison of LMMs trained in different languages.

\xhdr{Fine-grained classification} 
We categorize our dataset across several dimensions, including difficulty, grade level, and knowledge point. 
Problems are classified by difficulty---simple, medium, and hard---based on the average accuracy achieved by students. Simple problems have a scoring rate above 85\%, medium between 70\% and 85\%, and hard below 70\%. 
From Table~\ref{tab:CARP}, it can be seen that the number of problems of each difficulty level follows a Gaussian distribution.

Next, we organize questions by educational grade: seven, eight, and nine grade, representing the three years of junior middle school in China. 
Since higher-grade knowledge generally requires an understanding of lower-grade knowledge as a prerequisite, this classification allows us to better study whether the LMMs exhibit a similar dependency on prior knowledge when solving problems.

Additionally, each problem is tagged with specific knowledge points, identified based on insights from teachers. This enables targeted retrieval, application, and analysis of the model's knowledge gaps in specific areas. 
In Figure~\ref{fig:process_pie}, we present the knowledge point taxonomy and the proportion of data in each category.

%% file: text/003evaluation.tex
\section{Evaluation}
\label{evaluation methods}

\begin{figure*}[!ht]
    \centering
    \includegraphics[width=\textwidth]{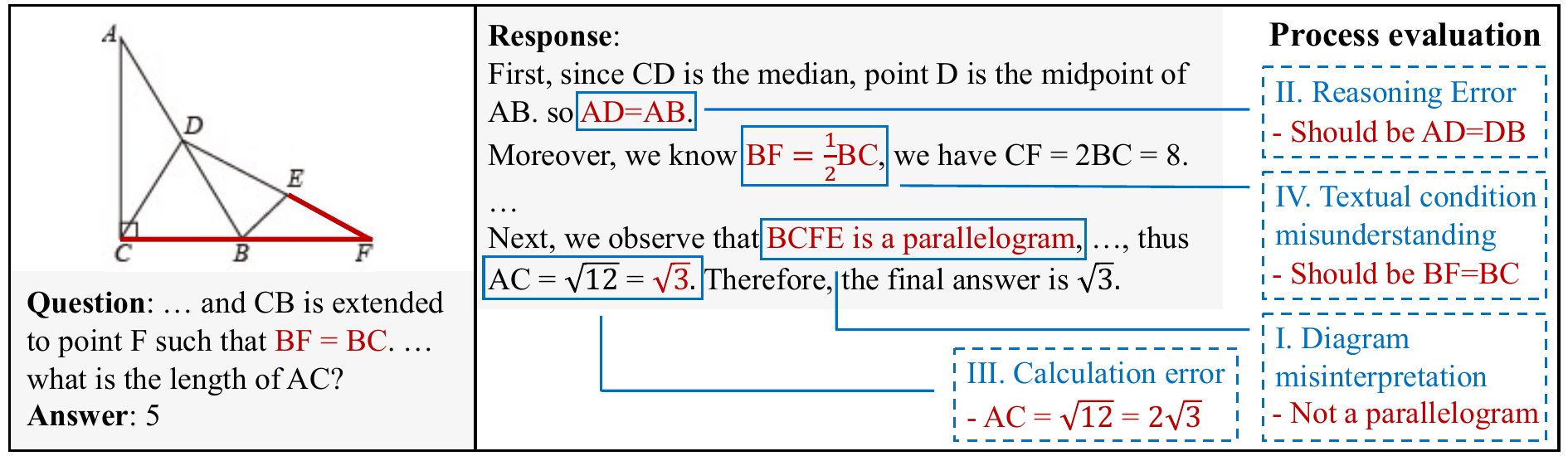}
    \caption{Example for four different types of errors in multimodal math reasoning.}
    \label{fig:error}
\end{figure*}

\subsection{Evaluation Protocols}

Recent advancements in LMMs have enabled the generation of textual responses for mathematical problem-solving~\cite{chen2024far,liu2024visual,hong2023cogagent,qi2024cogcom}, a process that imitates human reasoning in mathematics.
This capability introduces new evaluation criteria focusing on the generative nature of LMMs, especially concerning the intermediate solving steps.
Accordingly, we propose a systematic method for assessing the performance of LMMs in the MM-MATH datasets in Figure~\ref{fig:overview}, divided into three phases: (1) LMM generates formatted solutions to math problems, (2) Compare the generated solution against the groundtruth solution, and (3) Score the result to evaluate model performance and identify process errors.
Specifically, we input the textual problem and associated images, prompting the model to generate solutions with answers encapsulated in \textbackslash boxed\{\}. Our prompt details are provided in Appendix~\ref{prompt}. During the outcome comparison, we extract the final results from \textbackslash boxed\{\}. For process comparison, we use GPT-4V~\cite{openai2023gpt4v} to automatically perform a comparative analysis of the model-generated solutions against the groundtruth solutions.

\subsection{Evaluation Strategy}

Existing large model math benchmarks~\cite{he2024olympiadbench,wang2024measuring,liu2024mathbench} predominantly use binary comparison to assess the problem-solving capabilities of LLMs or LMMs, focusing on the final answer as the primary indicator of the model's capability. 
However, the accuracy of an answer typically relies on a correct intermediate reasoning process, with accurate reasoning steps leading to correct answers, and incorrect reasoning leading to erroneous solutions. 
By analyzing the model's solution process, we can identify the causes of errors and provide a more accurate assessment.
Thus, our evaluation incorporates two methods: outcome evaluation and process evaluation, designed to assess both the final answer and the model's reasoning process.

\xhdr{Outcome Evaluation}

Our evaluation requires that answers generated by LMMs be encapsulated within \textbackslash boxed\{\}, enabling direct comparison. 
We judge final answers according to their category: 
(1) For numerical answers, we accept the model's answers as long as the numerical gap to the groundtruth answer falls within a permissible error margin, e.g., 1.414 is acceptable for an answer of \(\sqrt{2}\) as their difference is less than 0.01. 
(2) For expression-type answers such as \(y = ax + b\), we utilize the SymPy package to simplify expressions. We then compare the model's simplified output with the groundtruth expression for exact matching.
(3) For interval-type answers like \((a, b)\) or \(a < x < b\), we standardize them into the format \((a, b)\) and verify the equality of boundary values. Additionally, we address special cases where models append extra signs to final results (e.g., cm) or generate exponential values like \(2^{2024}\), by removing the extra sign and transforming the values for proper comparison. 
We manually verified 500 evaluation results using our outcome evaluation pipeline and found only 13 errors.

\xhdr{Process Evaluation}
The problem-solving process of the multimodal model involves multiple factors, including a deep understanding of the problem conditions, extracting information from diagrams, and utilizing the models' knowledge to derive results. Consequently, our process evaluation takes the original textual question, associated image, and the groundtruth solution, and uses GPT-4V to compare the content generated by LMMs, with the prompt shown in Appendix~\ref{prompt}. 
The solutions generated by LMMs may contain numerous errors. In our prompt design, we aim to identify the \textbf{first error} in the model's generated process compared to groundtruth, since it is often the initial error that leads to further mistakes, resulting in incorrect outcomes. 
We use this first error to classify the cause of error in our process evaluation.
Through deeper examination, we find that the first identified error may sometimes not be the main error of the models' solution, which we will analyze further in Appendix~\ref{first error}. 
We classify the errors into four types, exemplified in Figure~\ref{fig:error}.

\noindent
I. \emph{Diagram misinterpretation}: This refers to the LMM’s inability to accurately understand the elements and their attributes in diagrams, such as the shapes, geometries, and their spatial relationships.

\noindent
II. \emph{Reasoning error}: This occurs when the model lacks or incorrectly applies logical reasoning knowledge. For instance, in the case of Figure~\ref{fig:error}, the model incorrectly reasons that $AD=AB$ from $D$ is the midpoint of $AB$, while $AD=DB$ should be the correct deduction.

\noindent
III. \emph{Calculation error}: This error arises from the computational step during problem-solving and includes mistakes caused by miscalculations in equations and functions.

\noindent
IV. \emph{Textual condition misunderstanding}: This type of error involves a model misinterpreting the given conditions of a textual problem. For example, in Figure~\ref{fig:error}, the problem states that $BF = BC$, but the model mistakenly interprets this condition as $BF = \frac{1}{2}BC$ during the solution process.

\begin{table*}[!htbp]
  \centering
  \small
  \caption{The outcome performance of both closed-source and open-source large models on MM-MATH in comparison with the human-level baseline. The evaluation involves three dimensions: \textit{difficulty}, \textit{grade levels}, and \textit{knowledge points}, each comprised of three fine-grained classes. The results are presented as percentages of accuracy.}
    \begin{tabular}[\linewidth]{l|rrr|rrr|rrr|r}
    \toprule
   \textbf{Model}  & \textbf{Easy} & \textbf{Medium} & \textbf{Hard} & \textbf{Seven} & \textbf{Eight} & \textbf{Nine} & \textbf{Trans} & \textbf{Shape} & \textbf{Func} & \textbf{Average} \\
    \midrule
    \multicolumn{11}{c}{\emph{Baseline}} \\
    \midrule
    \textbf{Human} & 90.7 & 81.9 & 47.6 & 85.6 & 73.7 & 77.9 & 81.1 & 83.2 & 77.5 & 80.4 \\
    \midrule
    \multicolumn{11}{c}{\emph{Large Multimodal Models (w/o Image)}} \\
    \midrule
    \textbf{Gemini-Pro-V} & 10.1 & 5.7 & 1.8 & 10.0   & 5.3 & 6.7 & 6.6 & 5.7 & 6.4 & 6.2 \\
    \textbf{Claude-3-Opus} & 31.7 & 17.3 & 7.2 & 32.5 & 14.9 & 2.2 & 20.8 & 18.5 & 12.9 & 19.2 \\
    \textbf{GPT-4} & 37.0  & 20.3 & 7.2 & 38.7 & 17.1 & 26.2 & 23.3 & 21.4 & 18.1 & 22.5 \\
    \textbf{GPT-4V} & 35.2 & 18.1 & 7.2 & 31.2 & 17.2 & 22.3 & 18.4 & 21.4 & 13.3 & 20.4 \\
    \textbf{GPT-4o} & 41.4 & 23.9 & 3.6 & 35.0 & 23.9 & 30.5 & 22.8 & 29.7 & 19.4 & 27.6 \\
    \midrule
    \multicolumn{11}{c}{\emph{Large Multimodal Models (w/ Image)}} \\
    \midrule
    \textbf{DeepSeek-VL-7B-Chat} & 17.4 & 4.7 & 1.4 & 7.5 & 6.6 & 3.9 &3.4 & 6.0 & 3.5 & 5.4 \\
    \textbf{Yi-34B-Chat} & 12.9 & 5.0  & 1.5 & 21.3 & 5.6 & 3.5 & 5.0 & 7.6 & 3.8 & 6.5 \\
    \textbf{LLaVA-V1.6-34B} & 8.8 & 5.4 & 1.8 & 12.6 & 6.5 & 4.2 & 4.0 & 6.5 & 3.8 & 5.8 \\
    \textbf{InternVL-4B-Chat-1.5} & 18.5 & 10.7 & 1.8 & 12.5 & 11.1 & 11.9 & 11.4 & 12.3 & 5.5 & 11.6 \\
    \textbf{Qwen-VL-Max} & 14.5 & 11.2 & 3.6 & 16.2 & 1.1  & 11.3 & 11.0 & 12.5 & 10.5 & 11.4 \\
    \textbf{Gemini-Pro-V} & 19.3 & 8.2 & 0.0     & 1.5  & 7.4 & 11.5 & 10.4 & 10.6 & 7.1 & 9.7 \\
    \textbf{Claude-3-Opus} & 29.5 & 19.3 & 3.6 & 32.5 & 16.4 & 23.0 & 20.6 & 21.7 & 16.9 & 20.3 \\
    \textbf{GPT-4V} & 37.8 & 21.2 & 1.8 & 28.7 & 17.9 & 28.0 & 22.2 & 24.7 & 19.5 & 23.1 \\
    \textbf{GPT-4o} & \textbf{45.8} & \textbf{30.0} & \textbf{10.9} & \textbf{40.0} & \textbf{26.0} & \textbf{36.0} & \textbf{30.7} & \textbf{33.7} & \textbf{26.2} & \textbf{31.8} \\
    \bottomrule
    \end{tabular}%

  \label{tab:exp}%
\end{table*}%

\section{Experiments}
\label{experimental results}

\subsection{Experimental Setup}
To comprehensively investigate the challenges of MM-MATH and the mathematical proficiency of models, we structure our experiments around two setups: (1) Text-Only Reasoning and (2) Multimodal Reasoning. For the first setting, we evaluate LMMs, including Gemini-Pro-V~\cite{team2023gemini}, Claude-3-Opus~\cite{anthropic2024claude3}, GPT-4\footnote{We use the \texttt{gpt-4-0125-preview} version for GPT-4.}~\cite{achiam2023gpt}, GPT-4V~\cite{openai2023gpt4v}, and GPT-4o~\cite{gpt4o} by providing only the textual contexts (\emph{i.e.,} questions) as inputs. For the second setting, we feed the entire multimodal contexts (\emph{i.e.,} questions and images) as inputs and evaluate both closed-source LMMs including GPT-4V, GPT-4o, Claude-3-Opus, Qwen-VL-Max~\cite{bai2023qwen} and open-source LMMs for DeepSeek-VL-7B-Chat~\cite{lu2024deepseek}, Yi-34B-Chat~\cite{young2024yi}, InternVL-4B-Chat-V1.5~\cite{chen2024far}, and LLaVA-V1.6-34B~\cite{liu2024llavanext}.

All selected models are capable of generating responses in the expected format, thus ensuring the validity of the evaluation.

\subsection{Outcome Evaluation Results}
We first analyze the performance of all models on the final outcomes of MM-MATH in comparison to a human-level baseline (the average performance of middle-school examinees from the online platform). The experimental results are shown in Table~\ref{tab:exp}. Here are our main findings from the results.

\paragraph{MM-MATH presents substantial challenges for current LMMs} From the evaluation results, we find that the most representative closed-source model to date, GPT-4o, performed the best across the board, achieving an average accuracy of 31.8\%, which significantly outperformed the best open-source model, InternVL-4B-Chat-1.5, with an average accuracy of 11.6\%. However, compared to the human-level baseline of 80.4\%, this best performance of the LMM still remains substantial room for improvement by 48.6\%.

\paragraph{LMMs gain limited benefits from visual contexts} Another notable observation is that LMMs with the text-only setups (\emph{i.e.,} only questions as inputs) exhibit only slight degradation in performance compared to the multimodal setups (\emph{i.e.,} questions and images as inputs). For example, there are differences of 4.2\%, 2.7\%, and 0.8\% for the models GPT-4o, GPT-4v, and Claude-3-Opus, respectively. This result suggests that current LMMs primarily rely on linguistic knowledge to solve mathematical problems, and their utilization of visual contexts is limited. Detailed case studies are provided in Appendix~\ref{text reasoning}.
\begin{figure*}[!ht]
    \centering
    \begin{minipage}[b]{0.58\textwidth}
        \centering
        \includegraphics[width=\textwidth]{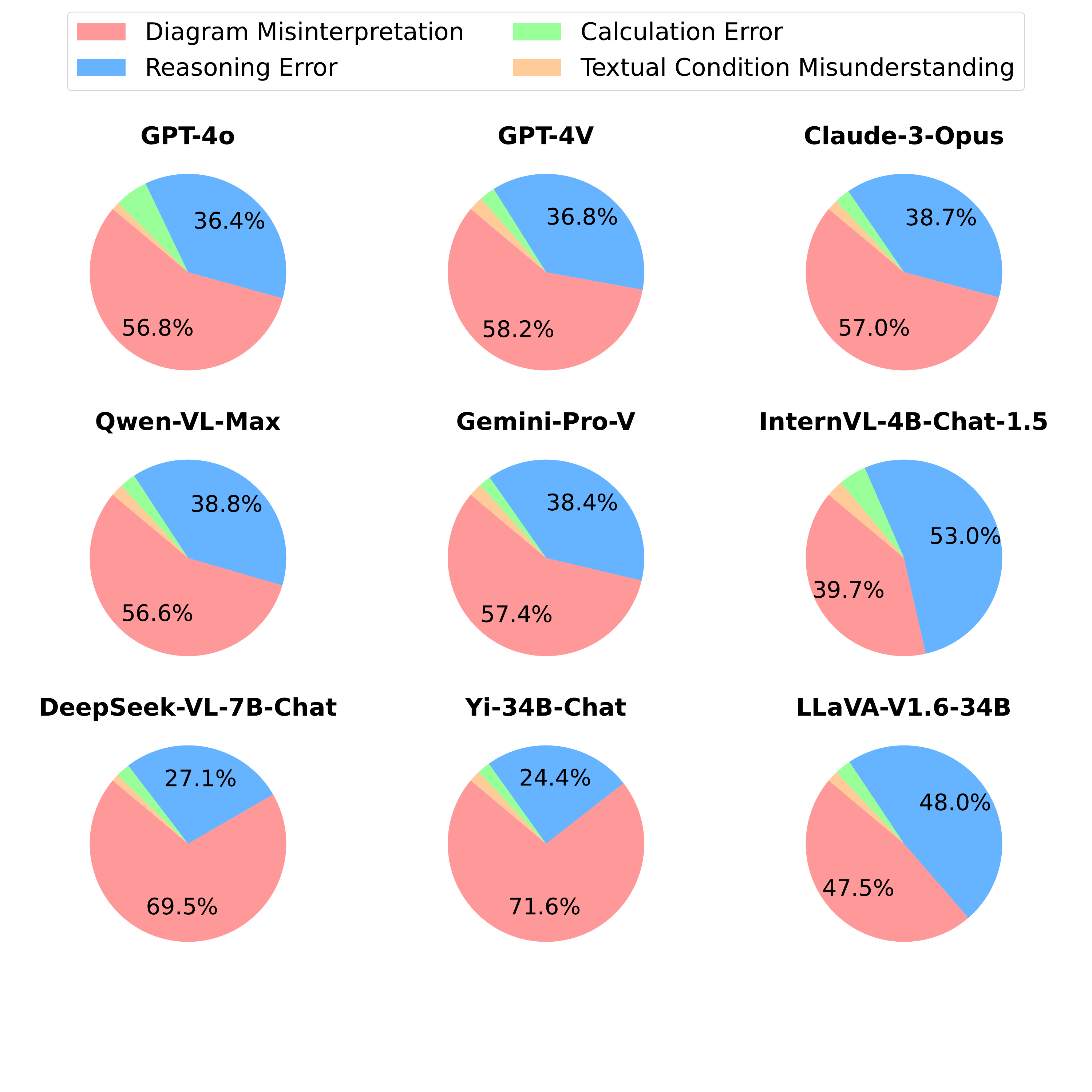}
        \caption{Proportion of four types of errors in various LMMs, with diagram misinterpretation errors and reasoning errors constituting the majority.} 
        \label{fig:pie}
    \end{minipage}
    \quad
    \begin{minipage}[b]{0.39\textwidth}
        \centering
        \includegraphics[width=\textwidth]{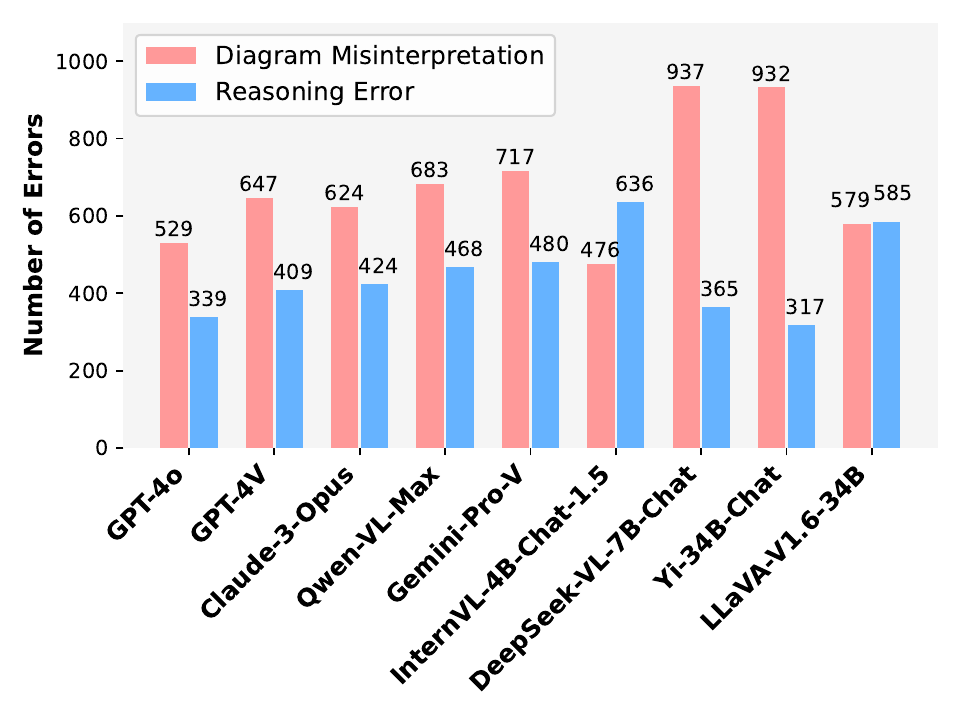}
        \caption{Number of the first two errors in evaluated LMMs.}
        \label{fig:image_error}
        % \vspace{0.5cm}
        \includegraphics[width=\textwidth]{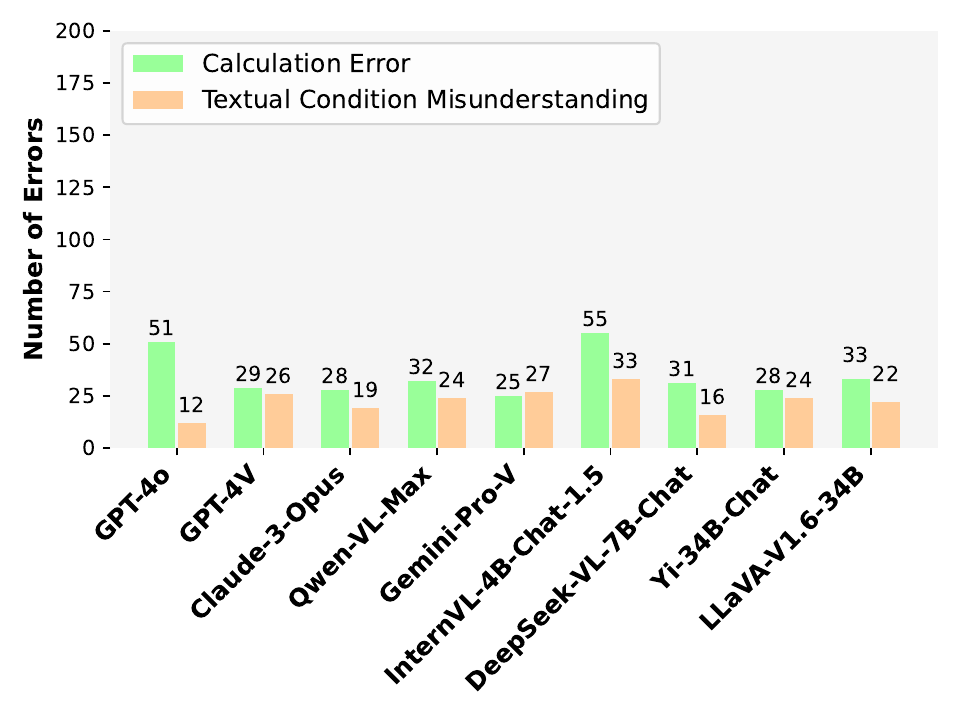}
        \caption{Number of the last two errors in evaluated LMMs.}
        \label{fig:image3}
    \end{minipage}
    \label{fig:three_images}
\end{figure*}

\paragraph{Conclusion from discriminative evaluation dimensions and capability distribution} In the difficulty dimension of MM-MATH, we can see the discriminative stepwise degradations in models performance on progressively challenging subsets (\emph{e.g.,} the 10.9\%, 30.0\%, and 45.8\% accuracy scores on \textit{Easy}, \textit{Medium} and \textit{Hard} subsets for GPT-4o). This result indicates that the proposed evaluation dimensions exhibit a significant differentiation across three difficulty levels, making it more beneficial for exploring the capability shortcomings of models. In addition, the three types of knowledge points also provide us with opportunities to understand the capabilities of models from different fine-grained perspectives.

Regarding the evaluation results on different grade levels, one notable finding is that the accuracy distribution of most models across the three grade levels is similar to the distribution of human behaviors. For example, the models GPT-4o, GPT-4V, and Claude-3-Opus all showed the best performance on the Seventh-grade subset (with 40.0\%, 28.7\%, and 32.5\% accuracy scores), followed by the ninth-grade subset (with 36.0\%, 28.0\% and 23.0\% accuracy scores), and were the least accurate on the seventh-grade subset (with 26.0\%, 17.9\% and 16.4\% accuracy scores, respectively). This result suggests that the learning curve of LMMs in solving mathematical problems is similar to that of humans but falls short of reaching the human cognitive level.

\subsection{Process Evaluation Results}

Benefiting from the comprehensive annotation, we further evaluate the models performance on the solution process to thoroughly investigate the causes of errors and pinpoint the weaknesses of LMMs. Considering the variability in natural language expressions, we employ GPT-4V to compare the solutions generated by LMMs with the groundtruth solutions, and identify the first error in the solutions to analyze the causes of errors. We empirically find that this method can effectively align the solutions from LMMs with the groundtruth, enabling an unbiased validation of the errors. Though effective, we find that there is still room for improvement in this measurement, with approximately 9\% of errors not being correctly identified (see the detailed analysis in Appendix~\ref{prompt_effect}).

Figure~\ref{fig:pie} illustrates the proportion of different error types in both open-source and closed-source LMMs. Figure~\ref{fig:image_error} and \ref{fig:image3} further show the number of errors for each error type.
Our main findings are detailed below.

\paragraph{Weak comprehension of elements in images is a major cause} It is evident that errors related to the recognition of image elements or their attributes constitute the highest proportion, exceeding half of the total errors.
This indicates that existing LMMs cannot yet sufficiently incorporate image information into their reasoning processes, limiting their efficacy in multimodal reasoning. Intriguingly, among closed-source LMMs---GPT-4o, GPT-4V, Claude-3-Opus, Gemini-Pro-V, and Qwen-VL-Max---the proportion of errors in image recognition are highly consistent, around 57\%. 
This might imply that the visual encoder modules used by these models have common issues and cannot handle certain types of images.
Additionally, the much lower proportion of diagram misinterpretation errors in InternVL-4B-Chat-1.5 (39.7\%) explains why a 4B small model has even better overall performance than Gemini-Pro-V (57.4\%) or Qwen-VL-Max (56.6\%).
Therefore, the key to enhancing the LMM's multimodal math problem-solving ability lies in understanding the visual context, and this step does not necessitate a large model size.
Examples of reasoning errors involving image elements and attributes are provided in Appendix~\ref{image_error}.

\paragraph{Multimodal models exhibit poor use of theorems during reasoning} We find that reasoning errors in large language models (LLMs) are often due to the incorrect application of theorems, accounting for about 40\% of overall errors. Misuse or omission of theorems misleads these LMMs, leading to errors (e.g., GPT-4V misuses the cosine rule, resulting in no solution, as detailed in Appendix~\ref{Misapplication of Theorems}). 
Unlike image understanding, we find that a larger model size effectively helps reduce reasoning errors in the model.
For instance, while InternVL-4B-Chat-1.5 exhibits fewer image understanding errors even with smaller model size, it still encounters more reasoning errors (636) compared to larger models such as Gemini-Pro-V (480) and Qwen-VL-Max (468).

\paragraph{Calculation is not a primary issue but reflects a capability gap} In the process evaluation of LMMs, calculation errors constitute a relatively lower proportion. However, the error in some models (e.g., GPT-4o, 51 errors) is significantly higher compared to others (e.g., GPT-4V, 29 errors). 
This indicates that while calculation is not the primary problem, equipping them with more powerful numerical computation capabilities can further boost the models' problem-solving success rates.

\paragraph{Models have an effective understanding of the textual problem} As shown in Figure~\ref{fig:pie}, among all nine models from both open-source and closed-source, the proportion of errors due to misunderstanding of the textual conditions is extremely small (less than 2\% of the total errors). This suggests that the text-based capabilities of LMMs are not the bottleneck in solving multimodal mathematical problems. Instead, we should focus more on fine-grained recognition and reasoning of visual content to enhance the capabilities of LMMs.

%% file: text/004related_work.tex
\section{Related Work}
Using large models to solve mathematical problems has recently become a research hotspot. GSM8k~\cite{cobbe2021training} has widely been used to evaluate the mathematical abilities of various LLMs~\cite{touvron2023llama,anil2023palm,gao2023pal}. However, its problems are relatively simple, and many models can achieve an accuracy rate of 90\% or higher. Recently, more challenging mathematical benchmarks~\cite{hendrycks2021measuring,liu2024mathbench,he2024olympiadbench} have emerged to further advance mathematical reasoning in language models, but these are typically text-only based reasoning.

Multimodal mathematical benchmarks trace back to the study of geometry problems~\cite{seo2015solving,chen2022unigeo}, where geometric elements are described through a specialized parsing language~\cite{seo2015solving,zhang2022plane,hao2022pgdp5k} or text described language~\cite{gao2023g}. Recent rapid developments in LMMs~\cite{alayrac2022flamingo,wang2023cogvlm,liu2024visual,qi2024cogcom} have led to numerous multimodal math benchmarks~\cite{lu2023mathvista, yue2024mmmu,ying2024mmt} to assess their capabilities. However, these benchmarks primarily composed of multiple-choice questions, evaluating model performance based on outcome examination. Given the dual nature of multimodal models---integrating both images and text---such simplistic evaluations are inadequate. Although some benchmarks, like MathVerse~\cite{zhang2024mathverse}, have begun to focus on the problem-solving process, they still rely on a binary evaluation approach. 
In comparison, our MM-MATH benchmark is constructed with step-by-step solution which enables both outcome and process evaluations of LMMs.

%% file: text/005conclusion.tex
\section{Conclusion}

This paper introduces MM-MATH, a challenging benchmark for evaluating multimodal math reasoning in LMMs. 
Our findings reveal while current LMMs demonstrate some reasoning ability, they heavily rely on textual information and struggle to utilize visual cues. This is evidenced by the minimal accuracy difference between text-only and multimodal settings, and the prevalence of diagram misinterpretation errors.
MM-MATH's fine-grained classification highlights the need for models that can handle varying problem difficulties and leverage knowledge across different grade levels.

\section{Limitations}

We limit our benchmark's mathematical knowledge to the middle school level, representing only a portion of K-12 education. In the future, we plan to expand the scope of MM-MATH to include high school and college-level multimodal mathematics.
Our evaluation results highlight the current deficiencies of LMMs in solving mathematical problems. While improvements to LMMs have not yet been made to address these shortcomings, our next step involves targeted training to enhance the models' problem-solving capabilities. We believe our dataset will significantly aid this process, as they contain detailed solutions paired with each problem.

%% file: text/006appendix.tex
\section{Data Source for Human Performance}
The 21st Century Education Network provides academic proficiency reports that analyze students' knowledge mastery after each exam. We compile the end-of-term exam scores for each problem. 
% as each grade level performance presented in Table~\ref{tab:exp}, to compare with the LMMs. Figure~\ref{fig:xueqing} shows the analysis of the end-of-term exam proficiency report for ninth-grade students, where the grade-level score is the average of the overall scores from each class.
% \begin{figure*}[!ht]
%     \centering
%     \includegraphics[width=0.8\linewidth]{image/xueqing.pdf}
%     \caption{The grade nine scores in Table~\ref{tab:exp} are compiled from the average end-of-term exam scores of each class.}
%     \label{fig:xueqing}
% \end{figure*}

\section{Open-Ended Transformation}
\begin{figure*}[!ht]
    \centering
    \includegraphics[width=0.8\linewidth]{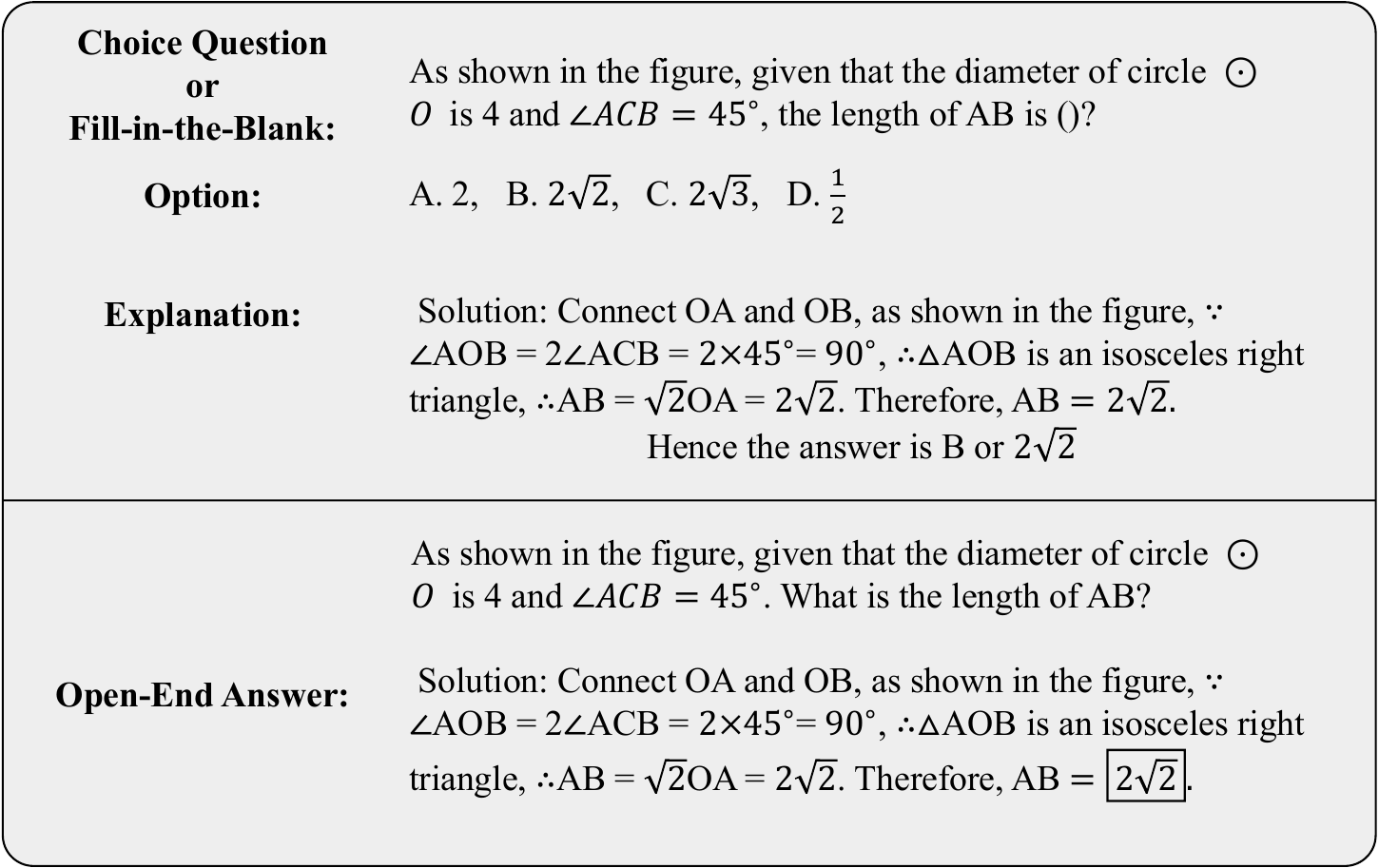}
    \caption{ An example of converting multiple-choice and fill-in-the-blank questions to open-ended format. The final part of the textual problem ``()'' is rewritten in descriptive language, and the main content of the explanation is used as the answer.}
    \label{fig:selection_trans}
\end{figure*}

\begin{figure*}[!ht]
    \centering
    \includegraphics[width=0.8\textwidth]{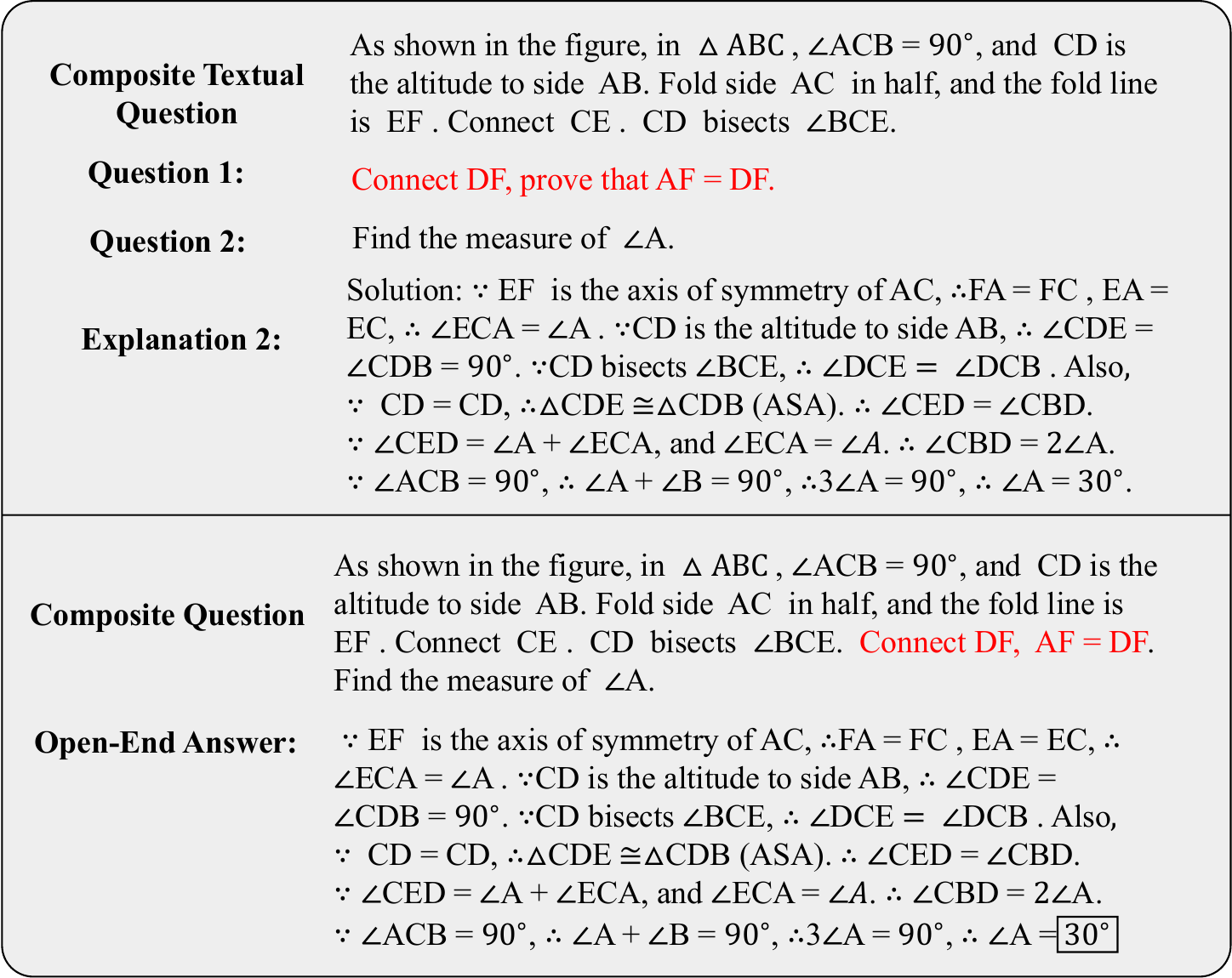}
    \caption{An example of converting composite question to open-ended format. Since Question 1 is a proof, we exclude it. We treat the main stem  as the stem of Question 2, and incorporate the conclusion of Question 1 (highlighted in red) as a new condition into the stem of Question 2.}
    \label{fig:composite_trans}
\end{figure*}

Our initial collection of MM-MATH problems includes four types: multiple-choice, fill-in-the-blank, open-ended, and composite questions. For multiple-choice and fill-in-the-blank questions, which include an answer and a step-by-step solution, we modifies the final part of the questions into descriptive language, removing single choice or fill-in-the-blank answers, and using the step-by-step solution as the answer described in Figure~\ref{fig:selection_trans}. For composite questions, we treat the main textual problem as the common stem for sub-questions and used the conclusion of one sub-question as the textual problem for the next described in Figure~\ref{fig:composite_trans}.
\label{open_end transformation}

\section{Prompt Design}
\label{prompt}
Table~\ref{tab:prompt} details the construction of the two types of prompts. For process evaluation prompts, our repeated experiments highlighted several key points: 1) Use the term ``\textit{incorrect}'' for \textit{textual condition misunderstanding}  to help GPT-4V classify the errors accurately. 2) Use the term ``\textit{misinterpretation}'' for \textit{diagram misinterpretation} errors to identify recognition mistakes during comparisons. 3) For reasoning errors, it is important to include specific examples.

For prompts that instruct the model to generate answers, we ensure the model produces a final answer enclosed in \textbackslash boxed\{\}.

\begin{table*}
    \centering
    \caption{This table presents the prompts used for process evaluation and answer generation by various LMMs in the MM-MATH benchmark.}
    \begin{tabular}{p{0.15\textwidth}p{0.2\textwidth}p{0.6\textwidth}}
    \toprule
    \textbf{Phase}      &       \textbf{Input}                            & \textbf{Prompt} \\
    \midrule
    \multirow{34}{*}{\makecell[l]{\bf Process\\\bf Evaluation\vspace{0.1cm}\\(GPT-4V)}}  &\multirow{34}{*}{\makecell[l]{Model's response\\Question\\Diagram\\Groundtruth Answer}} &Based on the given question stem, the diagram, and the correct answer, compare the model's response to identify the first error in model’s response. Then determine which of the following categories the error belongs to, or if there is no error, classify it as category five:\vspace{0.1cm}

1. Misinterpretation of diagram elements or properties: For example, incorrect coordinate recognition, identifying parallel lines as intersecting lines, or inventing or misusing elements or properties not present in the diagram (e.g., identifying a shape as a square when it is not).\vspace{0.1cm}

2. Incorrect application of math theorems: For instance, wrongly applying a specific theorem, such as using the Pythagorean theorem on a non-right triangle, or omitting necessary theorems, such as failing to apply the similarity theorem to obviously similar triangles.\vspace{0.1cm}

3. Calculation errors: Such as mistakes in addition, subtraction, multiplication, division, or square root calculations.\vspace{0.1cm}

4. Incorrect use of given question stems: For example, if the stem states AB=1/2CD but the model generates AB=CD, indicating a failure to use the condition correctly.\vspace{0.1cm}

5. Other: No errors.\vspace{0.1cm}

Provide a detailed analysis, including the first mistake, the reason for the classification, and the correct approach to solving the problem. If there are no errors, only provide the analysis. The output format should be:\vspace{0.1cm}

--\textbf{First error}: 

--\textbf{Error category}: 

--\textbf{Detailed analysis}: \\
\midrule

\multirow{5}{*}{\makecell[l]{\bf Answer\\\bf Generation\\\  (LMMs)\vspace{0.1cm}}}  &\multirow{5}{*}{\makecell[l]{Question\\Diagram}} & Solve the following mathematics problem, write out the solution process according to the question, and use the same LaTeX format as the question in the solution process. Please display the final answer in the format  \textbackslash boxed\{\}.

\\
    \bottomrule
    \end{tabular}
    \label{tab:prompt}
\end{table*}

\section{First Error Identified}
\label{first error}

\begin{figure*}[!ht]
    \centering
    \includegraphics[width=0.8\textwidth]{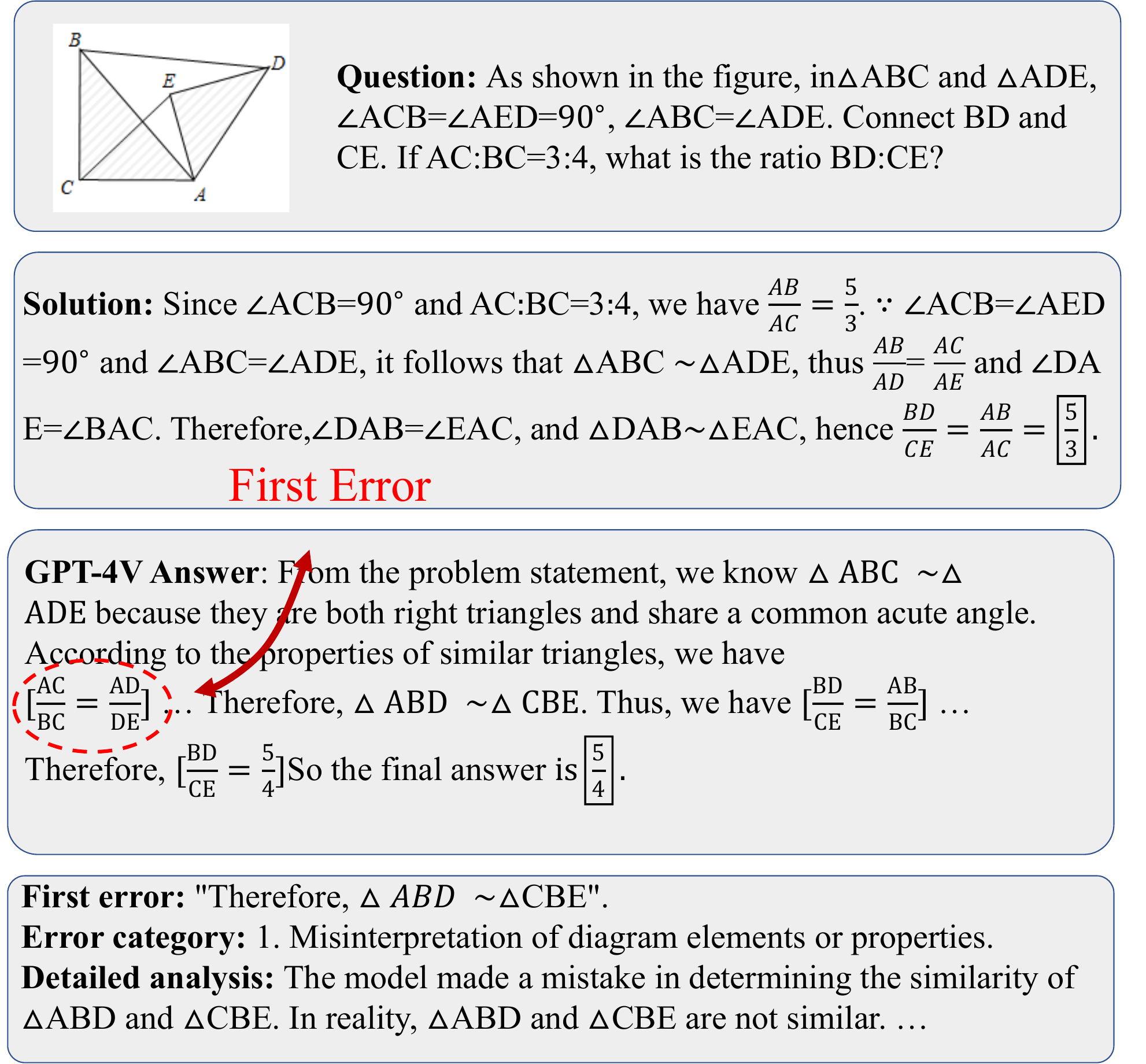}
    \caption{An example of first error detecting by GPT-4V}
    \label{fig:first_error}
\end{figure*}

The first error identified by GPT-4V, when comparing the problem-solving process generated by LMMs to the ground truth, may not necessarily be the initial error in the problem-solving process. As shown in Figure~\ref{fig:first_error}, the first error determined by GPT-4V is $\triangle ABD \sim \triangle CBE$ rather than intial error \[\frac{AC}{BC}=\frac{AD}{DE}.\]

\section{Text Reason First}
\label{text reasoning}
\begin{figure*}[!ht]
    \centering
    \includegraphics[width=0.8\textwidth]{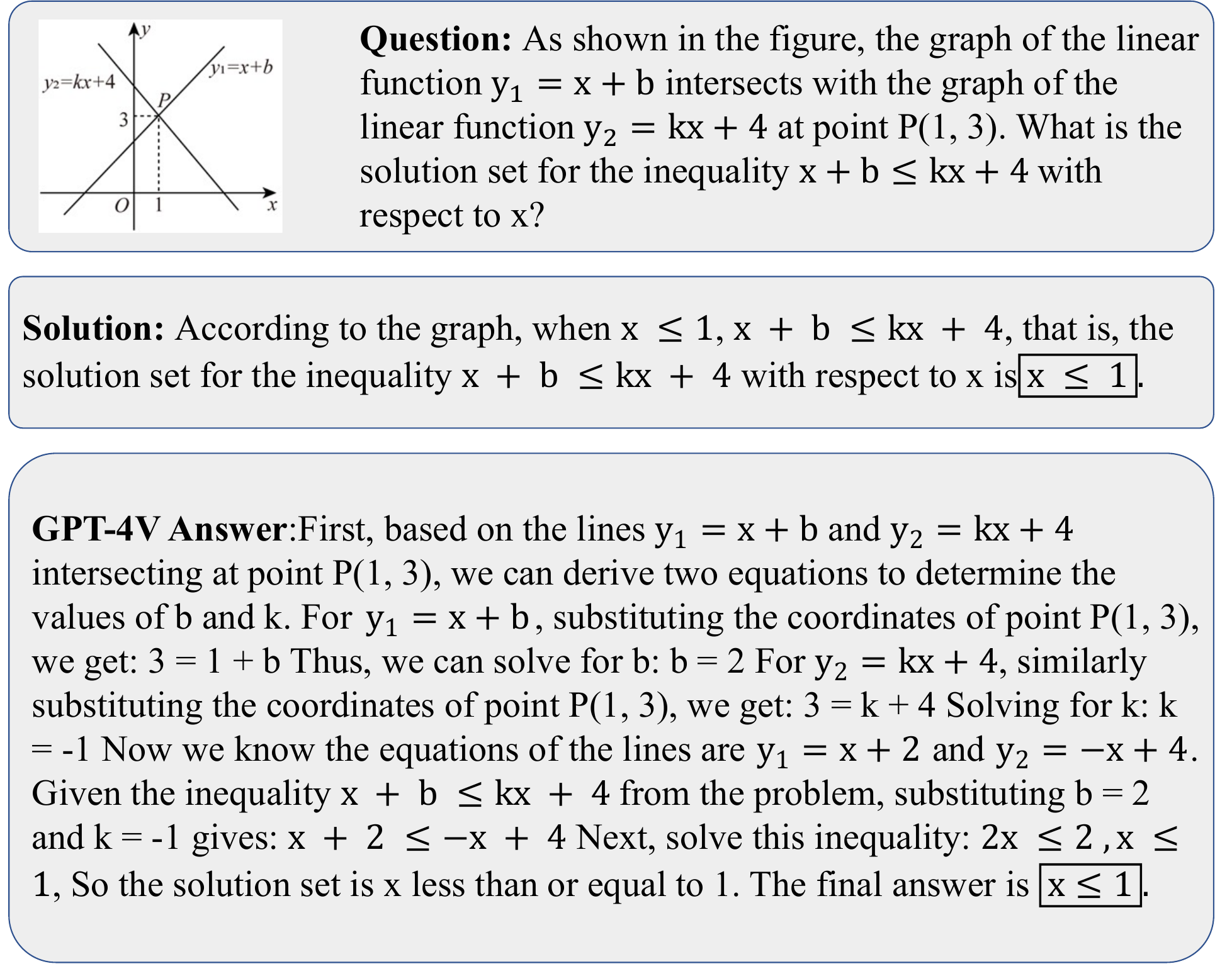}
    \caption{
A case where LMMs reasoning relies solely on textual analysis, resulting in increased complexity.}
    \label{fig:text_error1}
\end{figure*}

\begin{figure*}[!ht]
    \centering
    \includegraphics[width=0.8\textwidth]{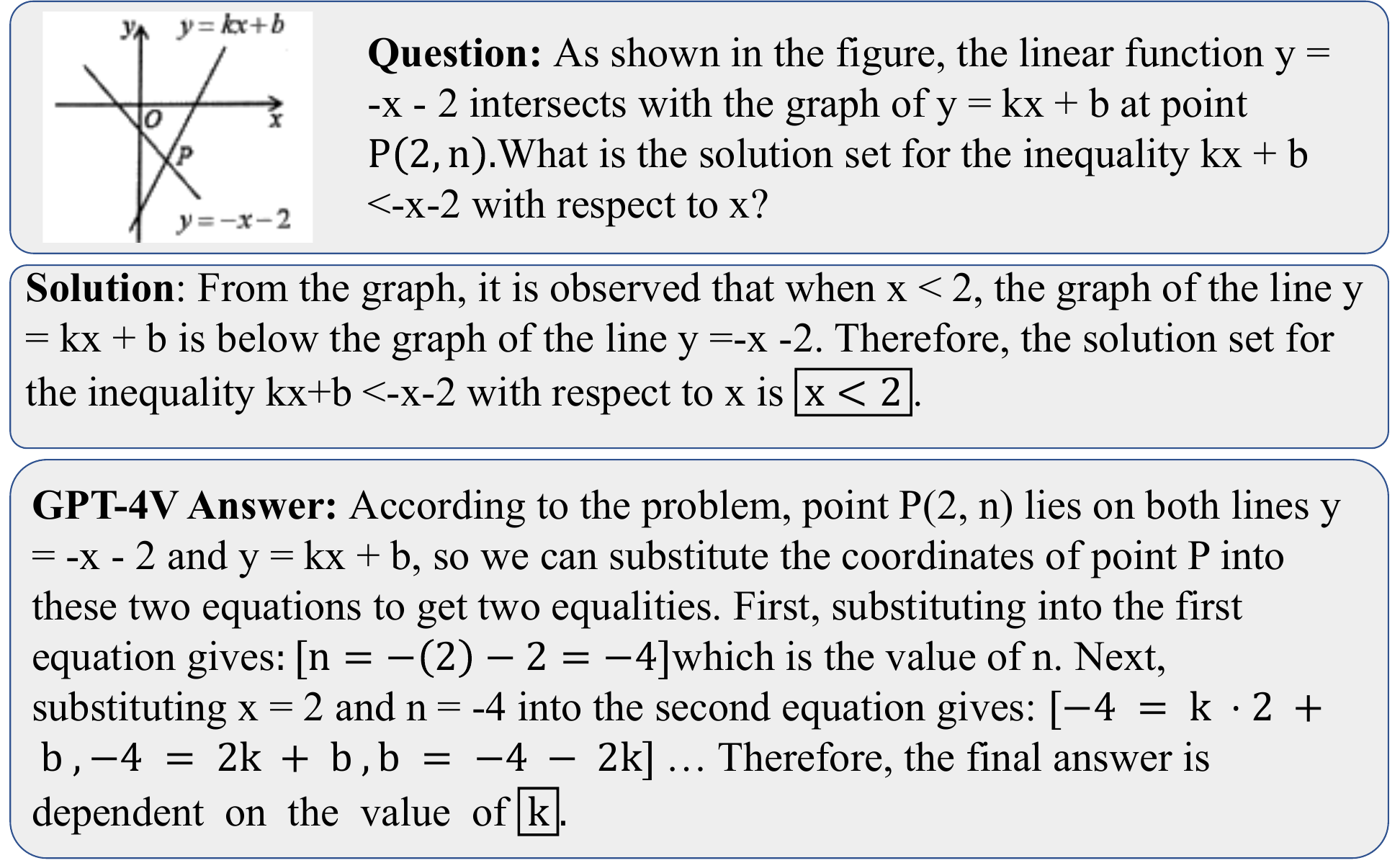}
    \caption{A case where LMMs reasoning ignores image information, relying only on text and leading to errors.}
    \label{fig:text_error2}
\end{figure*}

Figure~\ref{fig:text_error1} and Figure~\ref{fig:text_error2} illustrate examples of multimodal reasoning. Regardless of whether all problem conditions are provided, multimodal models tend to rely solely on textual analytical methods, neglecting the information in the images. This approach increases the complexity of problem-solving and leads to a higher likelihood of errors.

\section{Case of Prompt Effectiveness}
\label{prompt_effect}
\begin{figure*}[!ht]
    \centering
    \includegraphics[width=0.8\textwidth]{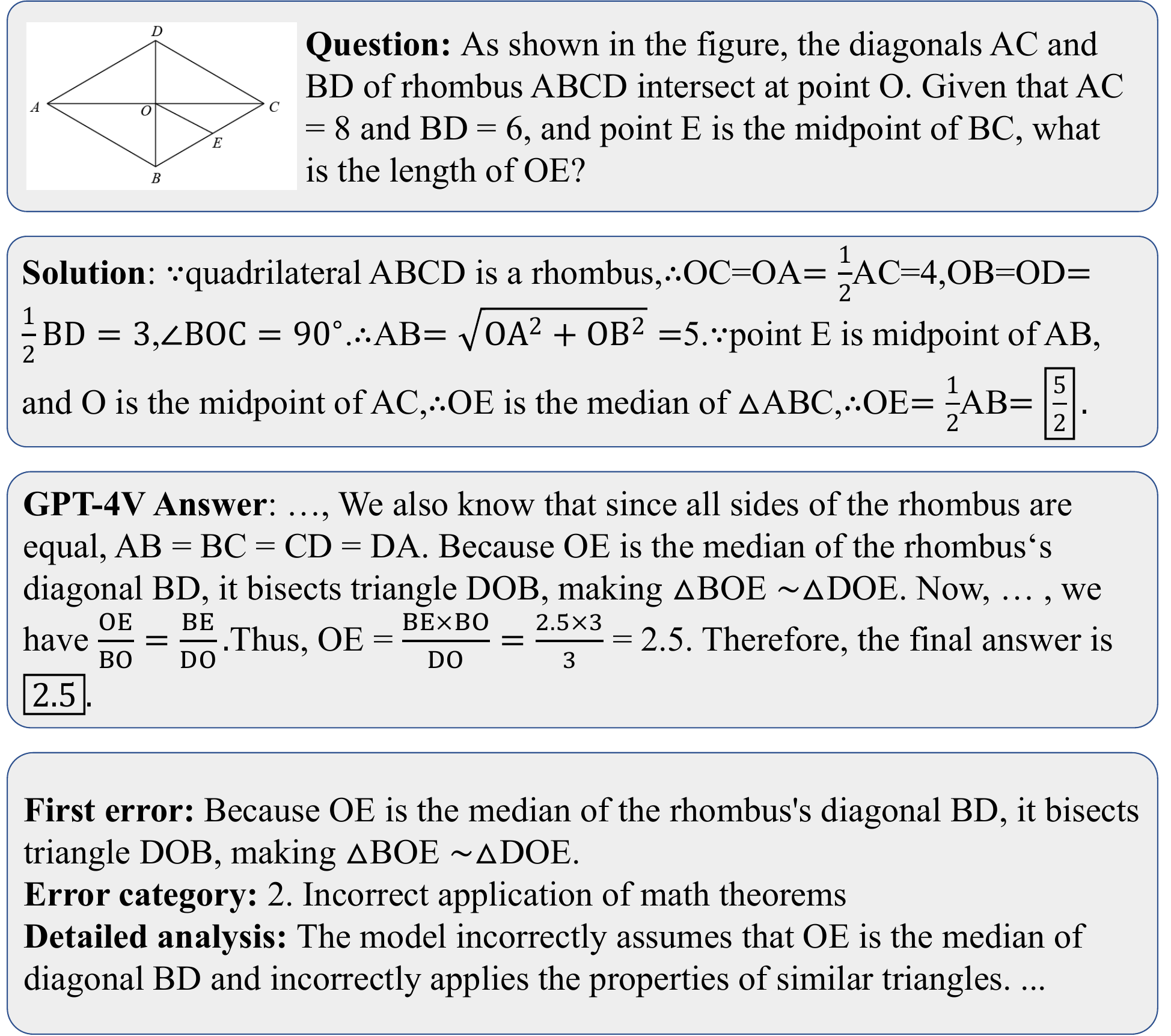}
    \caption{A case demonstrating the effectiveness of our prompt. GPT-4V successfully identified the reasoning error DOB, which was not present in the solution.}
    \label{fig:prompt_effect}
\end{figure*}

Figure~\ref{fig:prompt_effect} illustrates the effectiveness of our designed prompt. Although there is no textual description of DOB in the groundtruth problem-solving process, GPT-4V successfully identified and categorized the error. The error was classified as a mathematical theory error by GPT-4V because the solution did not apply the similarity theorem for reasoning. 

\section{Misapplication of Theorems}
\label{Misapplication of Theorems}
\begin{figure*}[!ht]
    \centering
    \includegraphics[width=0.8\textwidth]{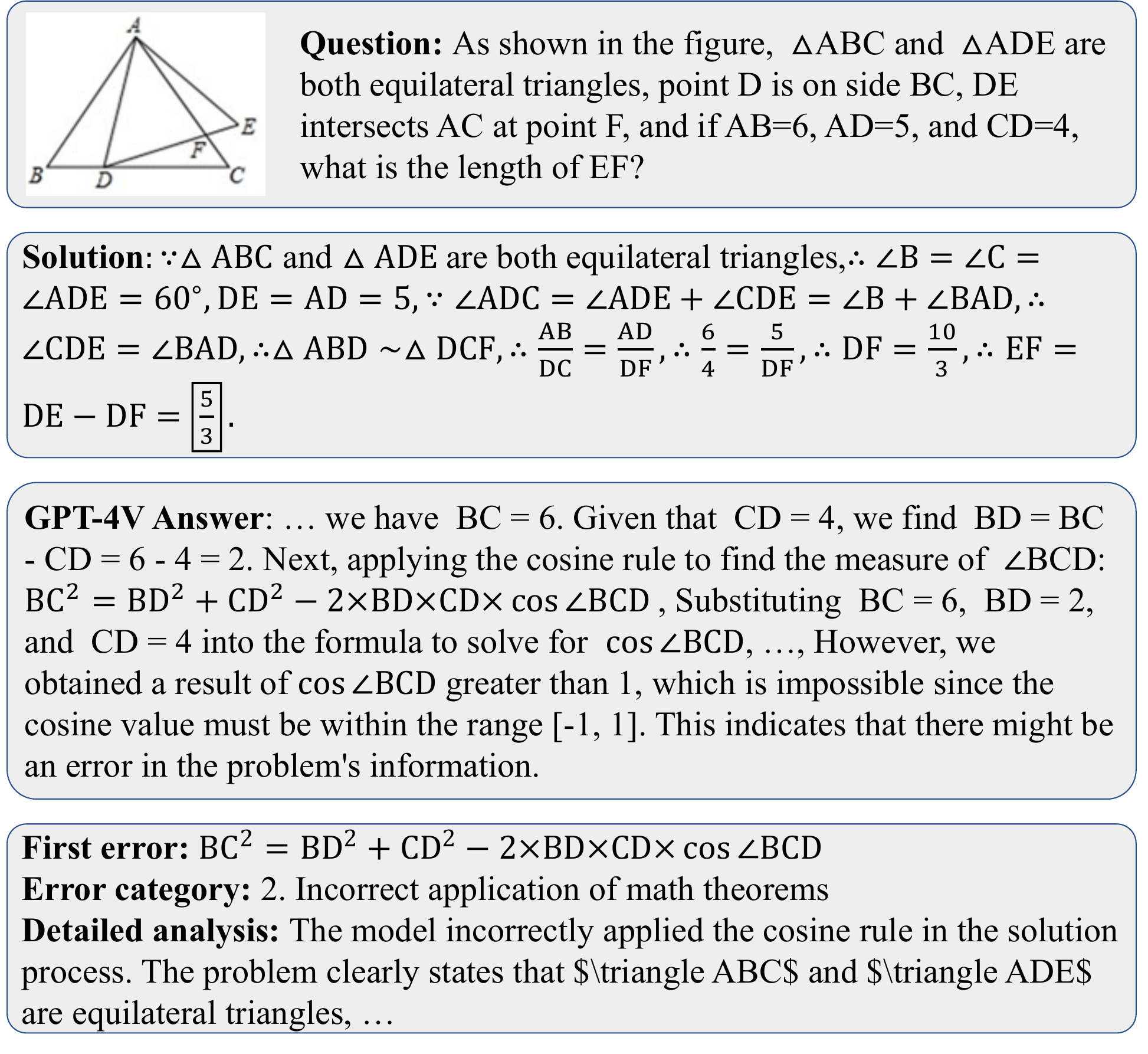}
    \caption{A reasoning error case: misuse of the cosine rule}
    \label{fig:theory_error}
\end{figure*}

Figure~\ref{fig:theory_error} illustrates a case where GPT-4V incorrectly applied the cosine theorem during problem-solving. In this case, the sides used in the cosine theorem do not form a triangle in the image. Despite this, the model forcibly applied the theorem, leading to calculation errors. This improper use of the theorem on incorrect image elements further resulted in erroneous problem-solving steps.

\section{Element Recognition Error in Image}
\label{image_error}
\begin{figure*}[!ht]
    \centering
    \includegraphics[width=0.8\textwidth]{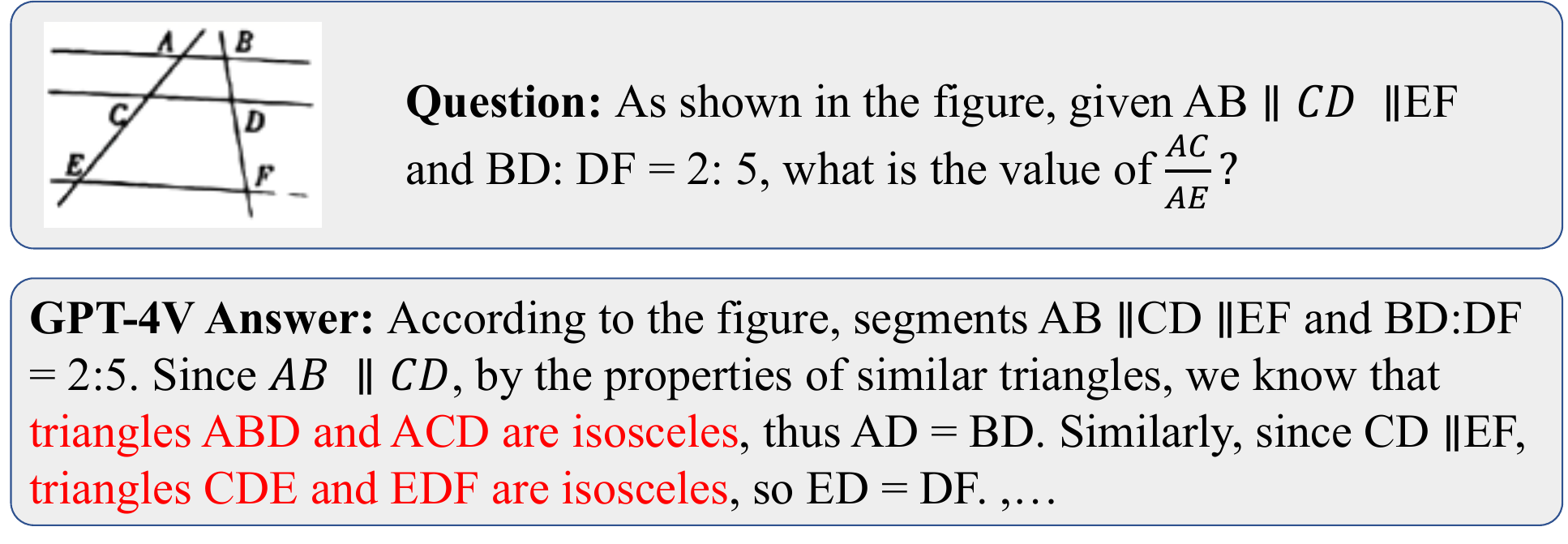}
    \caption{A case of diagram misinterpretation : parallel lines misidentified as a triangle.}
    \label{fig:image_case_error1}
\end{figure*}

\begin{figure*}[!ht]
    \centering
    \includegraphics[width=0.8\textwidth]{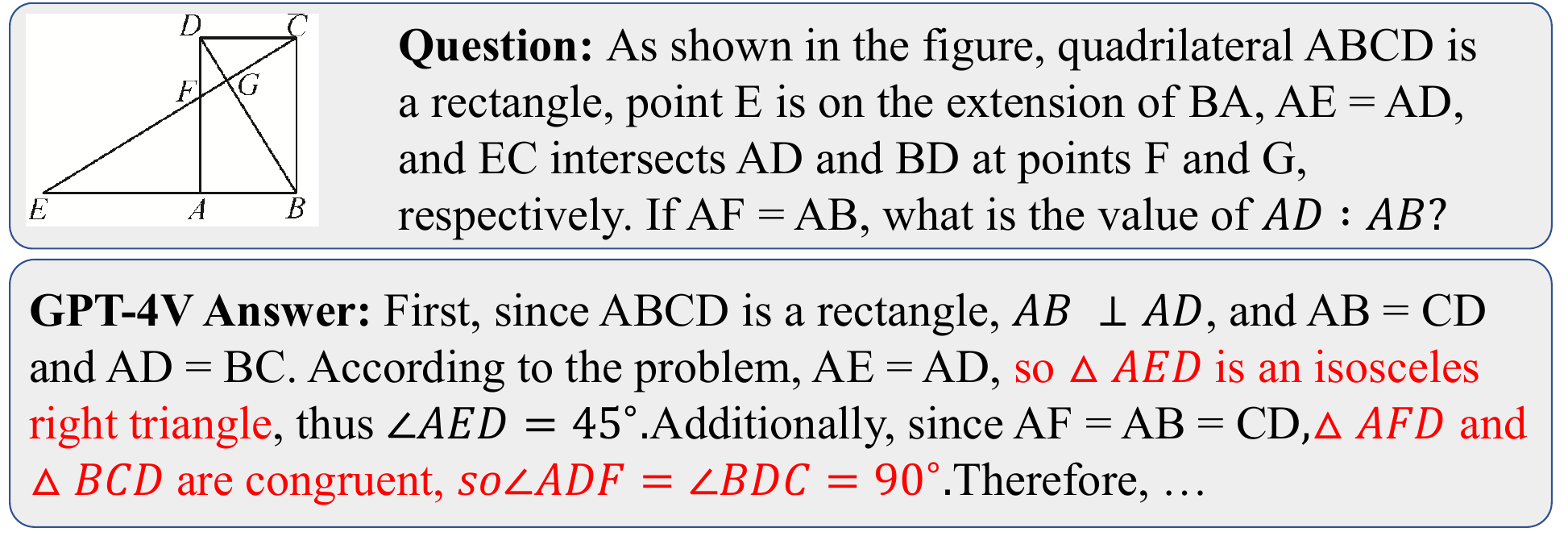}
    \caption{A case of diagram misinterpretation: a straight line misidentified as a triangle.}
    \label{fig:image_case_error2}
\end{figure*}

We present two cases where GPT-4V exhibited errors in image element recognition during problem-solving. In Figure~\ref{fig:image_case_error1}, the image contains only parallel lines; however, GPT-4V incorrectly identified a triangle. In Figure~\ref{fig:image_case_error2}, AFD is a straight line, but GPT-4V mistakenly perceived it as a triangle. These cases demonstrate GPT-4V's deficiencies in accurate image element recognition, leading to erroneous reasoning.

